# A resource-based Korean morphological annotation system

**Hyun-gue Huh**       **Éric Laporte**
Laboratoire d'informatique de l'Institut Gaspard-Monge
Université de Marne-la-Vallée/CNRS (UMR 8049)
5, bd Descartes – F77454 Marne-la-Vallée CEDEX 2 – France
`hhuh@univ-mlv.fr`     `eric.laporte@univ-mlv.fr`

## Abstract

We describe a resource-based method of morphological annotation of written Korean text. Korean is an agglutinative language. The output of our system is a graph of morphemes annotated with accurate linguistic information. The language resources used by the system can be easily updated, which allows users to control the evolution of the performances of the system. We show that morphological annotation of Korean text can be performed directly with a lexicon of words and without morphological rules.

## 1 Introduction

In the first phase of the processing of a written text, words are annotated with basic information, such as part of speech (POS). Two of the criteria of quality of a system performing this task are particularly relevant: the output of this process must be accurate and informative; the language resources used by the system must be able to undergo an evolution in a controlled way. The objective of the work reported here was to explore a method likely to enhance the performances of state-of-the-art systems of Korean morphological annotation in both regards. We decided to make language resources the central point of the problem.

Korean is one of the most studied agglutinative languages in the world. Korean words (*eojeol*) are delimited by spaces or other symbols, but a word usually consists of an undelimited concatenation of several morphemes: one or several stems, followed by zero, one or several functional morphemes. We will call stems all lexical morphemes, as opposed to functional or grammatical ones. The surface form of a morpheme occurring in a text may depend on neighbouring morphemes. Thus, it may differ from its base form or lexical form. These variations are termed as phonotactic, and the surface variants are called allomorphs. The objective of morphological annotation is to identify morphemes and assign relevant information to them.

The basic units of the Korean alphabet (*hangul*) are graphical syllables. These syllables can be decomposed over the Korean alphabet of simple and complex letters (*jamo*): each "letter" is either a cluster of 1 or 2 consonants, or a sequence of 1, 2 or 3 vowels. In addition, Chinese ideograms are sometimes employed also to write Korean stems.

It has been claimed that morphological annotation of Korean text could only be performed by a morphological analysis module accessing a lexicon of morphemes and using a set of rules. We show that it can also be performed directly with a lexicon of words. We describe an implemented system of morphological annotation that uses an actual lexicon of words. Our approach resorts to classical techniques of lexicon compression and lookup, but the application to an agglutinative language involved implementing new software components, which have received an open-source status.

In Section 2, we outline the state-of-the-art approach to Korean morphological annotation. Section 3 describes the alphabets and tag set used by our system. In Section 4, we explain our model of morpheme combinatorics. Section 5 reports the construction and use of the word lexicon. A conclusion and perspectives are presented in Section 6.

## 2 State of the art

Several morphological annotators of Korean text are available. The Hangul Analysis Module (HAM[1]) is one of the best Korean morphological analysers. Other fairly representative examples are described in (Shin et al., 1995; Park et al., 1998) and in (Lee et al., 1997a; Cha et al., 1998; Lee et al., 2002). The output for each morpheme is presented in two parts: the morpheme itself, and a grammatical tag. Morphemes are usually presented in their base form if they are stems, and in they surface form if they are functional morphemes. Tags are represented by symbols; they give the POS of stems, and grammatical information about functional morphemes. In output, 95% to 97% of morpheme/tag pairs are considered correct.

Morphological annotation of Korean text is usually performed in two steps (Sproat, 1992). In the first step, morpheme segmentation is performed with the aid of a lexicon of morphemes. This generates all possible ways of segmenting the input word. The second step makes a selection among the segmentations obtained and among the tags attached to the morphemes. The second step involves frequency-based learning from a tagged corpus with statistical models such as hidden Markov models, and sometimes also with error-driven learning of symbolic transformation rules (Brill, 1995; Lee et al., 1997a; Lee et al., 2002). Morphemes not found in the lexicon undergo a special treatment that guesses at their properties. A recent variant of this approach (Han and Palmer, 2005) swaps the main two steps: first, a sequence of tags is assigned to each word on the basis of a statistical model; then, morphological segmentation is performed with a lexicon of morphemes. The other approaches are less popular among searchers and language engineering companies. Some systems are based on two-level models, such as (Kim et al., 1994) and the Klex system of Han Na-rae[2]. (Choi, 1999) combines a lexicon of stems with a lexicon of endings with the aid of a connectivity table.

The delimitation of morphemes is provided, but some morpheme boundaries are usually modified so that they coincide with syllable boundaries. For example, if two suffixes make up a single syllable, like -셨:-*syôss*- which is a contraction of -으시:-*eusi*- (honorification towards sentence subject) and -었:-*ôss*- (past), they are usually considered as one morpheme. Such simplifications make it possible to encode morphemes on the Korean syllable-based alphabet, and are compatible with syllable-based models (Kang and Kim, 1994). However, they are an approximation.

We opted for the resource-based approach to obtain more accurate and more informative output.

The language resources used in annotators are corpora, rules and lexicons.

Corpus-based systems have an inherent lack of flexibility. A morphological annotator is not static infrastructure, it has to evolve with time. Due to the evolution of language across time, and especially of technical language, regular updates are necessary; a new application may involve the selection of a domain-specific vocabulary. The flexibility of a resource can be defined as the ability to control its evolution. In order to adapt a corpus-based system, one feeds a new corpus into the training process, since the operation of the system is dependent on the nature of the training corpus. A training process with a tagged corpus gives much better performance than unsupervised training (Merialdo, 1994). The extension of a system to input texts of new types or of a new period of time involves the costly task of tagging a corpus of new texts. Another type of evolution of a corpus-based system, a refinement of the tag set, such as the addition of new features, involves a re-tagging of existing tagged corpora, a task which is seldom achieved.

The situation is different with rules or lexicons. The flexibility of a manually constructed and updated rule set or lexicon depends on its level of readability and of non-redundancy (see section 4).

In current practice, words are segmented by a morphological analysis module that accesses a lexicon of morphemes and uses a set of rules. It has been claimed that morphological annotation of Korean text could only be performed this way, because a lexicon of words would be too large (e.g. Lee et al., 2002; Han and Palmer, 2005). We show that it can be performed directly with a lexicon of words; this solution dispenses with rules, thus simplifying and speeding up morphological annotation. The evidence given by Han and Palmer (2005) in support of their claim is the fact that the number of different words in Korean is very large, which is undisputed. In fact, they implicitly assume that the lexicon would be obtained by sequentially generating all words and associated information. Such a naive procedure would surely be impractical. Our system constructs a lexicon of words without generating any list of words at any of the phases of its construction or maintenance.

---

[1] http://nlp.kookmin.ac.kr/HAM/kor/ham-intr.html
[2] http://www.cis.upenn.edu/~nrh/klex.html

In our design, all morphological rules are applied to all possible configurations during the compilation of the resources and stored in a lexicon of words, which is searched during text annotation. No morphological rules are applied then. The lexicon of words occupies less than 600 Kb, and specifies 138,000,000 surface forms of words obtained from 39,130 base-form stems. The size of the lexicon does not grow with the number of words, due to our adaptation to Korean of state-of-the-art technology for lexicon management (Appel and Jacobson, 1988; Silberztein, 1991; Revuz, 1992; Lucchesi and Kowaltowski, 1993). Our approach could even be adapted further to allow for constructing a lexicon with infinitely many words.

## 3 Alphabets and tag set

Our system uses three Unicode character sets: the Korean syllabic alphabet, the Korean alphabet of letters, and the Chinese ideograms. The lexicon of words is constructed from a set of language resources that has been manually constructed and is manually updated by Korean linguists (Nam, 1996). In order to ensure that these resources are readable, they are encoded in the Korean syllabic alphabet. The only situation when this is impossible is when a morpheme boundary does not coincide with a syllable boundary. In that case, the morpheme boundary divides the syllable into two parts; one of these parts has no vowel and cannot be encoded in the syllabic alphabet: it is then encoded in the Korean alphabet of letters, which is another zone of the Unicode character set. This convention allows for an accurate delimitation of surface forms and base forms of all morphemes, including irregular ones. Chinese ideograms are provided in the information on Sino-Korean stems, which are sometimes spelled in Chinese ideograms in texts.

In the lexicon of words itself, words are encoded over the Korean alphabet of letters, for more efficient lexicon search. During text annotation, words in the text are converted into letters before the lexicon is searched.

Our tag set is more fine-grained than state-of-the-art: it comprises 173 tags for stems [to be compared to 18 in Lee et al. (2002) and 14 in Han and Palmer (2005)], and 84 tags for functional morphemes [15 in Lee et al. (2002) and in Han and Palmer (2005)]. Tags are more informative.

In addition, the tags are structured. They combine a general tag taken in a list of 16 general tags, and 0 to 4 features specifying subcategories. The list of general tags is displayed in Table 1. There are 91 features with a total of 152 values.

| A | adjective | Sfx | derivational suffix |
|---|---|---|---|
| ADV | adverb | Morph | pre-final verb/adj. ending |
| DET | determiner | Post | postposition |
| N | noun | Sc | conjunctive suffix |
| NI | bound noun | Sd | determinative suffix |
| PRO | pronoun | Sncomp | nominalization suffix |
| V | verb | St | final ending |
| INT | interjection | Suf | pre-final nominal ending |

**Table 1.** General tags

This structure is in conformity with emerging international standards in representation of lexical tags (Lee et al., 2004). Tag sets in previous Korean morphological analysers were unstructured or hierarchical (Lee et al., 2002), not feature-based.

## 4 Morphotactics and connectivity

The final part of some verbal and adjectival stems undergoes phonotactic variations when a suffix is appended to them. For example, the stem 크:*keu-* "big" becomes ㅋ:*k-* before the suffix -었:-*ôss-* (past). In order to reduce the level of redundancy of manually updated resources, lexicons of base-form stems were con-structed. Each stem was assigned a structured tag. Stem allomorphs are generated from base-form stems with 71 transducers of the same type as those used to inflect words in inflectional languages (Silberztein, 2000). The input part of the transducer specifies letters to remove or to add in order to obtain the allomorph from the base form. The output part specifies the tag and compatibility symbol (see below) to be assigned to the allomorph. These transducers are viewed and edited in graphical form with the open-source Unitex system[3] (Paumier, 2002).

The combination of a stem with a sequence of suffixes obeys a number of constraints. Checking these constraints is necessary to discard wrong segmentations. We distinguish two types of suffixes: derivational and inflectional.

Derivational suffixes are markers of verbalization, adjectivalization and adverbialization. They are appended by applying transducers of the same type as above. In our current version, 8 transducers append derivational suffixes. These transducers invoke 5 subgraphs, thus constituting recursive transition networks (RTN).

Inflectional suffixes comprise all other types of suffixes. A single (possibly derived) stem can be combined with up to 5,500 different sequences of inflectional suffixes. Compatibility between stems and inflectional suffixes is represented by a set of 59 compatibility symbols (CS). Each stem and stem allomorph is assigned a CS, which defines the set of suffix sequences that can be appended to it. The CSs take into account two types of constraints: grammatical and phonotactic constraints. CSs are comparable with adjacency symbols, except that they include the constraints between all the morphemes in a word, not only between adjacent morphemes. They convey more information than adjacency symbols, but they are less numerous: 59 to be compared to 300 (Lee et al., 2002). The lexicon of stems assigns CSs to base stems. CSs are automatically assigned to stem allomorphs during the generation of allomorphs.

Connectivity between suffixes obeys phonotactic and grammatical constraints. Phonotactic constraints affect surface forms, whereas grammatical constraints affect base form/tag pairs. The standard model for representing both types of constraints is the finite-state model. For example, Lee et al. (2002) use a table that encodes connectivity between morphemes with the aid of morpheme tags and adjacency symbols. Such a table can be viewed as a finite-state automaton in which the states are the adjacency symbols and the transitions are labelled by the morpheme tags. In Kim et al. (1994) and in the Klex system of Han Na-rae, these constraints are represented in the two-level formalism, which is equivalent to regular expressions, which are in turn equivalent to finite-state automata. All these forms are computationally relevant, but they are little readable: the inclusion of a new item or the correction of an error is error-prone. Two-level rules have a very low level of redundancy, but they are complex to read because they combine a morphological part and a logical part (the symbols <=>, <=, =>).

In our system, connectivity constraints between suffixes are represented in finite-state transducers, i.e. finite-state automata with input/output labels. These transducers describe sequences of suffixes. Their input represents surface forms and their output represents base forms and tags. We introduced two innovations in order to enhance their readability. Firstly, they are edited and viewed graphically. Secondly, since most of the transducers are large and would not display conveniently on a single screen or page, they take the form of RTNs: transitions can be labelled by a call to a sub-transducer instead of an input/output pair. The 59 CSs correspond to 59 transducers. Most of the sub-transducers that they call are shared, which reduces the level of redundancy of the system. The total number of simple graphs making up the RTNs is 230.

In the case of several of the RTNs, the graph of calls to sub-transducers admits cycles. Due to these cycles, these RTNs generate an infinite set of endings. The lexicon compiler allows for keeping the set of generated endings finite by breaking all cycles.

## 5 Word lexicon

The various readable resources described above are compiled into an operational lexicon of words whenever one of them is updated. The lexicon of words has an index for fast matching. This index is a finite-state transducer over the Korean alphabet of letters. This is a transposi-tion of the state-of-the-art technology of representation of lexicons of forms in inflectional languages (Appel and Jacobson, 1988; Silberztein, 1991; Revuz, 1992; Lucchesi and Kowaltowski, 1993). Another index structure, the trie, has been tested with the same lexicon. The size of the trie (930 Kb) is slightly larger than the size of the transducer (560 Kb), due to the representation of endings which is repeated many times in the trie.

---

[3] http://www-igm.univ-mlv.fr/~unitex/manuelunitex.pdf

The compilation of the lexicon of words from the readable resources follows several sequential steps. First, all resources are converted from the Korean syllabic alphabet to the Korean alphabet of letters. In a second step, lexicons of stem allomorphs and of derived stems are generated from the base-form stem lexicons by applying the transducers with Unitex. In a third step, the resulting lexicons of stems are compiled by the Unitex lexicon compiler. Each compiled lexicon has an index, which is a finite-state automaton. The final states of the automaton give access to the lexical information, and in particular to the CSs of the stems. In a fourth step, each transducer of sequences of suffixes is converted into a list by a path enumerator, and each of these lists is processed by the lexicon compiler. The names of the compiled ending lexicons contain the corresponding CSs. In the final step, the stem lexicons and the ending lexicons are merged into a word lexicon. This operation links the final states of the stem lexicons to the initial states of the corresponding ending lexicons. The path enumerator and the lexicon link editor have been implemented for this experiment and will receive an open-source status. The path enumerator allows for breaking cycles in the graph of calls to sub-transducers, so that the enumeration remains finite.

The current version of this compilation process generates a lexicon of one-stem words only. Multi-stem words will be represented in later versions.

These operations are independent of the text to be annotated; they are performed beforehand. They need to be repeated whenever one of the language resources is updated.

The operation of the morphological annotator is simple. The text is pre-processed for sentence segmentation, and tokenised (words are tokens). In each word, Korean syllables are converted into Korean letters; then, the lexicon of words is searched for the word. Lexicon search is efficient: it processes 41,222 words per second on a P4-400 Windows PC. When Chinese ideograms occur in a stem, the lexicon search module searches directly the lexical information attached to stem entries. We did not include any modules for processing words not found in the lexicon.

All analyses that are conform to phonotactic and grammatical in-word constraints are retained. However, checking these constraints does not suffice to remove all ambiguity from Korean words. A thorough removal of ambiguity requires a syntactic process (Voutilainen, 1995; Laporte, 2001). Our system presents its output in an acyclic finite-state automaton (also called a graph or a lattice) of morphemes, as in Lee et al. (1997b), but displayed graphically. The output for each morpheme is presented in three parts: surface form, base form, and a structured tag providing the general tag of Table 1 and syntactic features. Word separators such as spaces are also present in this automaton.

The annotation of an evaluation sample by the system presented 67 % recall and 46 % precision. The annotation of a morpheme was considered wrong when any of the features was wrong. Among these errors, 78 % are resource errors that can be corrected by updating the resources, whereas the correction of the remaining 22 % would involve enhancing the compilation procedure.

## 6  Conclusion

We experimented with a method likely to enhance the performances of state-of-the-art systems of Korean morphological annotation. We made language resources the central point of the problem. All complex operations were integrated among resource management operations. The output of our system is accurate and informative; the language resources used by the system can be easily updated, which allows users to control the evolution of the performances of the system. Morphological annotation of Korean text is performed directly with a lexicon of words and without morphological rules, which simplifies and speeds up the process.

This work opens several perspectives. The resources will be extended by running the annotator and analysing output. Existing approaches to the analysis of unrecognised morphemes could be combined to our system: such approaches are complementary to our resource-based approach, and would take advantage of the rich information provided on the neighbouring words. Ambiguity resolution techniques can be applied to the output of our annotator: the syntactic approach would take advantage of the rich linguistic information provided in output; classical statistical approaches and priority rules (Kang, 1999) are applicable as well. Finally, a parallel system is under construction for Finnish, another agglutinative language with undelimited morphemes.